%% file: aaai24.tex
\documentclass[letterpaper]{article} 
\usepackage{aaai24}  
\usepackage{times}  
\usepackage{helvet}  
\usepackage{courier}  
\usepackage[hyphens]{url}  
\usepackage{graphicx} 
\usepackage{natbib}  
\usepackage{caption} 
\usepackage{algorithm}
\usepackage{algorithmic}

\usepackage{newfloat}
\usepackage{listings}
\DeclareCaptionStyle{ruled}{labelfont=normalfont,labelsep=colon,strut=off} 
\floatstyle{ruled}
\newfloat{listing}{tb}{lst}{}
\floatname{listing}{Listing}
\usepackage{amsfonts}
\usepackage{amssymb}
\usepackage{bm}
\usepackage{booktabs}
\usepackage{multicol}
\usepackage{multirow}
\usepackage{adjustbox}
\usepackage{amsmath}

\DeclareMathOperator*{\argmin}{arg\,min}

\usepackage{cleveref}
\usepackage{textcomp}

\newcommand{\method}{\texttt{LEO}}
\newcommand{\smac}{{SMAC}}

\newcommand{\ml}{ML}
\newcommand{\smacI}{{SmacI}}
\newcommand{\smacD}{{SmacD}}
\newcommand{\minValByWt}{{MaxRatio}}
\newcommand{\minWt}{{MinWt}}
\newcommand{\lex}{{Lex}}

\newcommand{\evo}{EVO}
\newcommand{\vo}{VO}

\newcommand{\paretofront}{PF}

\title{\method: Learning Efficient Orderings for Multiobjective Binary Decision Diagrams}
\author{
    Rahul Patel,
    Elias B. Khalil    
}
\affiliations{
    Department of Mechanical and Industrial Engineering, University of Toronto\\
    rm.patel@mail.utoronto.ca, khalil@mie.utoronto.ca

}

\usepackage{bibentry}

\begin{document}
\nocopyright

\maketitle

\begin{abstract}
Approaches based on Binary decision diagrams (BDDs) have recently achieved state-of-the-art results for multiobjective integer programming problems. The variable ordering used in constructing BDDs can have a significant impact on their size and on the quality of bounds derived from relaxed or restricted BDDs for single-objective optimization problems. We first showcase a similar impact of variable ordering on the Pareto frontier (\paretofront{}) enumeration time for the multiobjective knapsack problem, suggesting the need for deriving variable ordering methods that improve the scalability of the multiobjective BDD approach. To that end, we derive a novel parameter configuration space based on variable scoring functions which are linear in a small set of interpretable and easy-to-compute variable features. We show how the configuration space can be efficiently explored using black-box optimization, circumventing the curse of dimensionality (in the number of variables and objectives), and finding good orderings that reduce the \paretofront{} enumeration time. However, black-box optimization approaches incur a computational overhead that outweighs the reduction in time due to good variable ordering. To alleviate this issue, we propose \method{}, a supervised learning approach for finding efficient variable orderings that reduce the enumeration time. Experiments on benchmark sets from the knapsack problem with 3-7 objectives and up to 80 variables show that~\method{} is $\sim$30-300\% and $\sim$10-200\% faster at \paretofront{} enumeration than common ordering strategies and algorithm configuration. Our code and instances are available at~\url{https://github.com/khalil-research/leo}.
\end{abstract}

\input{Sections/intro}
\input{Sections/prelim}

\input{Sections/related}

\input{Sections/method}
\input{Sections/study}

\input{Sections/results}

\input{Sections/conclusion}



\bibliography{aaai24}

\clearpage
\appendix
\section{Appendix}
\input{Sections/appendix}

\end{document}

%% file: Sections/intro.tex
\section{Introduction}
In many real-world scenarios, one must jointly optimize over a set of conflicting objectives. 
For instance, solving a portfolio optimization problem in finance requires simultaneously minimizing risk and maximizing return.
The field of multiobjective optimization deals with solving such problems.
It has been successfully applied in novel drug design \citep{lambrinidis2021multi}, space exploration \citep{song2018multi, tangpattanakul2012multi}, administrating radiotherapy  \citep{yu2000multi}, supply chain network design \cite{altiparmak2006genetic}, among others.
In this paper, we specifically focus on multiobjective integer programming (MOIP), which deals with solving multiobjective problems with integer variables and linear constraints.


The goal of solving a multiobjective problem is to find the Pareto frontier (\paretofront{}): the set of feasible solutions that are not dominated by any other solution, i.e., ones for which improving the  value of any objective deteriorates at least one other objective. The~\paretofront{} solutions provide the decision-maker with a set of trade-offs between the conflicting objectives. 
Objective-space search methods iteratively solve multiple related single-objective problems to enumerate the \paretofront{} but suffer from redundant computations in which previously found solutions are encountered again or a single-objective problem turns out infeasible.
On the other hand, decision-space search methods leverage branch-and-bound.
Unlike the single-objective case where one compares a single scalar bound (e.g., in mixed-integer linear programming (MIP)), one needs to compare bound \textit{sets} to decide if a node can be pruned; this in itself is quite challenging. 
Additionally, other crucial components of branch-and-bound such as branching variable selection and presolve are still underdeveloped, limiting the usability of this framework.


Binary decision diagrams (BDDs) have been a central tool in program verification and analysis~\cite{bryant1992symbolic, wegener2000branching}. More recently, however, they have been used to solve discrete optimization problems \citep{bergman2016decision, BergmanCHH16} that admit a recursive formulation akin to that of dynamic programming. 
BDDs leverage this structure to get an edge over MIP by efficiently encoding the feasible set into a network model which enables fast optimization.
To the best of our knowledge, \citet{bergman2016multiobjective} were the first to use BDDs to solve multiobjective problems, achieving state-of-the-art results for a number of problems. 
The \vo{} used to construct a BDD has a significant impact on its size and consequently any optimization of the diagram. 
However, the \vo{} problem within BDD-based MOIP has not been addressed in the literature.
We address this gap by designing a novel learning-based BDD \vo{} technique for faster enumeration of \paretofront{}.

\begin{figure*}[htbp!]
    \centering
    \includegraphics[width=\linewidth]{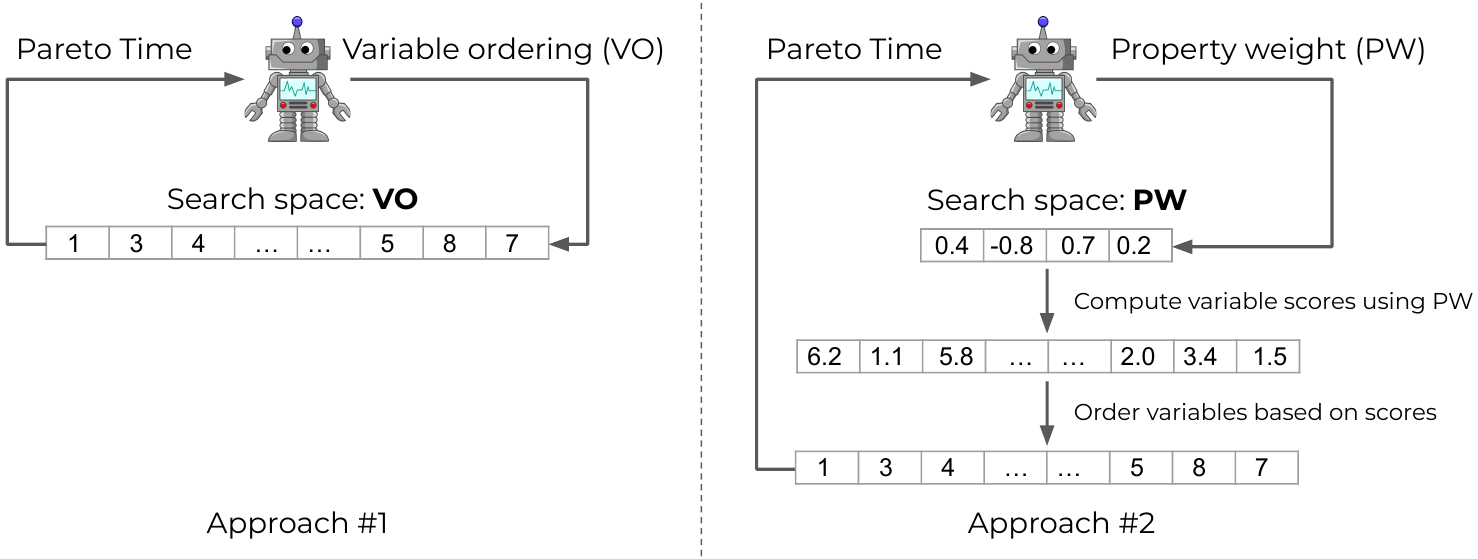}
    \caption{Searching for variable orderings. The robot represents any automated search procedure that tries to find a good variable ordering (Approach \#1) or property weights (Approach \#2) by iteratively proposing a variable ordering, measuring its run time on one or more instances, and repeating until a satisfactory ordering has been found. Approach \#1 depicts a naive approach that searches for an efficient variable ordering directly in the variable ordering space. 
    The proposed Approach \#2 conducts an indirect search by controlling a small set of variable property weights (PWs).}
    \label{fig:approaches}
\end{figure*}

We begin with the following hypothesis: \vo{} has an impact on the \paretofront{} enumeration time and an ``efficient" \vo{} can reduce it significantly. Following an empirical validation of this hypothesis, we show that such orderings can be found using black-box optimization, not directly in the (exponentially large) space of variable orderings, but rather indirectly in the space of constant-size variable scoring functions. The scoring function is a weighted linear combination of a fixed set of variable properties (or attributes), and the indirect search is in the space of possible weight combinations.
\Cref{fig:approaches} illustrates how a search in the property-weight space can alleviate the curse of dimensionality to make VO search scalable to problems with many variables.
However, solving the black-box optimization problem may be prohibitively time-consuming for any one instance. For variable ordering to be useful in practice, the time required to produce a good VO should be negligible relative to the actual PF enumeration time.
To alleviate this issue, we train a supervised machine learning (ML) model on the orderings collected using black-box optimization.
A trained model can then be used on unseen (test) instances to predict variable orderings. Should such a model generalize well, it would lead to reduced \paretofront{} enumeration times. We refer to our approach as~\method{} (Learning Efficient Orderings).
Our key contributions can be summarized as follows: 
\begin{enumerate}
    \item We show that variable ordering can have a dramatic impact on solving times through a case study of the multiobjective knapsack, a canonical combinatorial problem.
    \item We show how black-box optimization can be leveraged to find efficient variable orderings at scale.
    \item We design a supervised learning framework for predicting variable orderings which are obtained with black-box optimization on a set of training instances. Our ML models are invariant to permutations of variables and independent of the number of variables, enabling fast training and the use of one ML model across instances of different sizes. 
    \item We perform an extensive set of experiments on the knapsack problem and show that \method{} is  $\sim$30-300\% and $\sim$10-200\% faster than the best non-ML ordering strategy and the SMAC algorithm configuration tool, respectively.
    \item We perform a feature importance analysis of the best class of ML models we have found, extreme gradient boosted ranking trees. The analysis reveals that: (a) a single ML model can be trained across instances with varying numbers of objectives and variables; and (b) a single knapsack-specific feature that we had not initially contemplated performs reasonably well on its own, though far worse than our ML models.
\end{enumerate}

%% file: Sections/prelim.tex
\section{Preliminaries}
\label{sec:prelim}

\subsection{Multiobjective Optimization}
An MOIP problem takes the form $\mathcal M:= \min_{x} \left\{ \bar z(x): x \in \mathcal X, \mathcal X \subset \mathbb Z^n_+ \right\}$, where $x$ is the decision vector, $\mathcal X$ is a polyhedral feasible set, and $\bar z: \mathbb R^n \rightarrow \mathbb R^p$ a vector-valued objective function representing the $p$ objectives.
In this work, we focus on the knapsack problem with binary decision variables, hence $\mathcal X \subset \{0,1\}^n$.

\noindent
\textbf{Definition: Pareto dominance.}
Let $x^1, x^2 \in \mathcal X, \bar y^1 = \bar z(x^1), \bar y^2 = \bar z(x^2)$, then $\bar y^1$ \textit{dominates} $\bar y^2$ if $\bar y^1_j \leq \bar y^2_j, \forall j \in [p]$ and $\exists j\in[p]: \bar y^1_j < \bar y^2_j$. 

\noindent
\textbf{Definition: Efficient set.}
A solution $x^1 \in \mathcal X$ is called an efficient solution if $\nexists~x^2 \in \mathcal X$ such that $x^2$ dominates $x^1$.
The set of all efficient solutions to a multiobjective problem is called an efficient set $\mathcal X_E$.

\noindent
\textbf{Definition: Pareto frontier (\paretofront{}).}
The set of images of the efficient solutions in the objective space, i.e., $\mathcal Z_N = \{\bar z(x): x \in \mathcal X_E\}$, is called the \paretofront{}.

The exact solution approaches to solving multiobjective problems focus on efficiently enumerating its \paretofront{}.

\subsection{BDDs for Multiobjective Optimization}
A BDD is a compact encoding of the feasible set of a combinatorial optimization problem that exploits the recursive formulation of the problem.
Formally, a BDD is a layered acyclic graph $G=(n, \mathcal N, \mathcal A, \ell, d)$, composed of nodes in $\mathcal N$, arcs in $\mathcal A$, a node-level mapping $\ell: \mathcal N \rightarrow [n+1]$ that maps each node to a decision variable, and an arc-label mapping $d: \mathcal A \rightarrow \{0, 1\}$ that assigns a value from the variable domain of an arc's starting node to the arc. Here, $n$ is the number of variables of the multiobjective problem $\mathcal M$.

The nodes are partitioned into $n+1$ layers $L_1, \dots, L_{n+1}$, where $L_l=\{u: \ell(u) = l, u \in \mathcal N\}$.
The first and last layers have one node each, the root $\mathbf{r}$ and terminal nodes $\mathbf{t}$, respectively.
The width of a layer $L_l$ is equal to the number of nodes in that layer, $|L_l|$.
The width of a BDD $G$ is equal to the maximum layer width, $\max_{l \in [n+1]}|L_l|$.
An arc $a:=(r(a), t(a)) \in \mathcal A$ starts from the node $r(a) \in L_l$ and ends in node $t(a) \in L_{l+1}$ for some $l \in [n]$.
It has an associated label $d(a) \in \{0, 1\}$ and a vector of values $\bar v(a) \in \mathbb R^p_+$ that represents the contribution of that arc to the $p$ objective values. 

Let $\mathcal P$ represent all the paths from the root to the terminal node. 
A path $e(a_1, a_2, \dots, a_{n}) \in \mathcal P$ is equal to the solution $x=(d(a_1),d(a_2), \dots, d(a_n))$
and the corresponding image of this solution in the objective space is given by $\bar v(e) = \sum_{i=1}^{n} \bar v(a_i)$, where the sum is taken elementwise over the elements of each vector $\bar v(a_i)$.
The BDD representation of $\mathcal M$ is valid if $\mathcal Z_N = \texttt{ND}\left( \bigcup_{e \in \mathcal P} \bar v(e) \right)$, where $\mathcal Z_N$ is the \paretofront{} of $\mathcal M$ and $\texttt{ND}$ an operator to filter out dominated objective vectors. 
We refer the readers to \citet{bergman2021network} and \citet{bergman2016multiobjective} for the detailed description of the multiobjective knapsack BDD construction. In what follows, we assume access to a BDD construction and \paretofront{} enumeration procedure and will focus our attention on the variable ordering aspect.


%% file: Sections/related.tex
\section{Related Work}
\subsection{Exact Approaches for Solving MOIP Problems}
Traditional approaches to exactly solve MOIP can be divided into objective-space search and decision-space search methods. 
The objective-space search techniques \citep{kirlik2014new, boland2014triangle, boland2015criterion, ozlen2014multi} enumerate the \paretofront{} by searching in the space of objective function values. 
They transform a multiobjective problem into a single one either by weighted sum aggregation of objectives or transforming all but one objective into constraints.
The decision-space search approaches \citep{przybylski2017multi, sourd2008multiobjective, parragh2019branch, adelgren2022branch, belotti2013branch} instead explore the set of feasible decisions. 
Both these approaches have their own set of challenges as described in the introduction. 
We point the reader to \citep{ehrgott2016exact, ehrgott2006discussion} for a more detailed background. 

\citet{bergman2021network} showcased that BDD-based approaches can leverage problem structure and can be orders of magnitude faster on certain problem classes; the use of valid network pruning operations along with effective network compilation techniques were the prime factors behind their success. 
However, they do not study the effect of \vo{} on the \paretofront{} enumeration time.

\subsection{Variable Ordering for BDD Construction}
Finding a \vo{} that leads to a BDD with a minimal number of nodes is an NP-Complete problem \citep{bollig1996effect, bollig1996improving}.
This problem has received significant attention from the verification community as smaller BDDs are more efficient verifiers. 
Along with the size, \vo{} also affects the objective bounds; specifically, smaller exact BDDs are able to obtain better bounds on the corresponding limited width relaxed/restricted BDDs \citep{bergman2016decision}. 

The \vo{} techniques for BDD construction can be broadly categorized as  exact or heuristic. 
The exact approaches to \vo{} \citep{friedman1987finding, ebendt2007exact, ebendt2005combining, bergman2012variable}, though useful for smaller cases, are not able to scale with problem size. 
To alleviate the issue of scalability for larger problems, heuristic \vo{} techniques are used; the proposed methodology falls into this category. 
These heuristics can be general or problem-specific \citep{rice2008survey, bergman2016decision, lu2000efficient, aloul2003force, chung1993efficient, butler1991heuristics, rudell1993dynamic} but the literature has not tackled the multiobjective optimization setting.
A VO method can also be classified as either~\textit{static} or~\textit{dynamic}. Static orderings are specified in advance of BDD construction whereas dynamic orderings are derived incrementally as the BDD is constructed layer-by-layer. The latter may or may not be applicable depending on the BDD construction algorithm; see~\cite{karahalios2022variable} for a discussion of this distinction for Graph Coloring. We focus on static orderings and discuss extensions to the dynamic setting in the Conclusion. 

We focus on ML-based heuristics for \vo{} as they can be learned from a relevant dataset instead of handcrafted heuristics developed through a tedious trial-and-error process.

\subsection{Machine Learning for Variable Ordering}
\citet{grumberg2003learning} proposed a learning-based algorithm to construct smaller BDDs for model verification.
In particular, they create random orders for a given instance in the training set and tag variable pairs based on their impact on the resulting BDD size. 
Using this data, they learn multiple pair precedence classifiers. 
For a new instance at test time, they query each trained pair precedence classifier to construct a precedence table.
These tables are merged into one to derive the ordering. 
The success of this method hinges on the selective sampling of informative variable pairs and the ability to generate orders with sufficient variability in BDD size to increase the chance of observing high-quality labels. 
As they leverage problem-specific heuristics for selective sampling and rely on random sampling for generating labels, this method is not applicable to our setting. 
Specifically, we do not have a notion of how informative a given variable pair is. 
Additionally, random orders may not produce high variability in \paretofront{} enumeration time as evidenced by our experiments.

\citet{carbin2006learning} uses active learning to address the \vo{} problem for BDDs used in program analysis with the goal of minimizing the run time, similar to that of \method{}.
However, certain differences make it less amenable to our setting. 
Specifically, the technique to generate the candidate \vo{}s in \citet{carbin2006learning} is grounded in program analysis and cannot be applied to our problem. 
Instead, \method{} leverages bayesian optimization through \smac{} in conjunction with the novel property-weight search space to generate \vo{} candidates. 



\citet{drechsler1996learning} use an evolutionary search approach to learn a single variable ordering heuristic for a set of instances. The learned heuristic is a sequence of BDD operations (e.g., swapping two variables in an ordering) applied to instances from circuit design and verification, where the BDD represents a boolean function that must be verified efficiently. This approach can be seen as one of algorithm configuration, which we will compare to using SMAC and ultimately outperform.

The work of \citet{cappart2019improving} is the first learning-based method to address \vo{} problem for BDDs used in solving discrete optimization problems.
Specifically, they learn a policy to order variables of relaxed/restricted BDDs to obtain tighter bounds for the underlying optimization problem using reinforcement learning (RL). 
A key component of training an RL policy is devising the reward that the agent receives after taking an action in a given state and moving to the next state. 
Note that we care about reducing the time to enumerate the \paretofront{}, which is only available at the end of a training episode. 
The absence of intermediate rewards in our setting makes RL inapplicable.

\citet{karahalios2022variable} developed an algorithm portfolio approach to select the best \vo{} strategy from a set of alternatives when constructing relaxed decision diagrams for the (single-objective) graph coloring problem. 
Fundamental to a portfolio-based approach is the existence of a set of strategies, such that each one of them is better than the others at solving some problem instances.
However, such an algorithm portfolio does not exist in our case and part of the challenge is to discover good ordering strategies.

There have been some recent contributions  \cite{wu2022graph, liu2023end}  relating to solving multiobjective problems using deep learning and graph neural networks. However, these approaches are not exact and thus beyond the scope of this paper. We discuss potential extensions of~\method{} to the inexact setting in the Conclusion.

%% file: Sections/method.tex
\begin{figure*}[htbp!]
    \centering
    \includegraphics[width=0.8\linewidth]{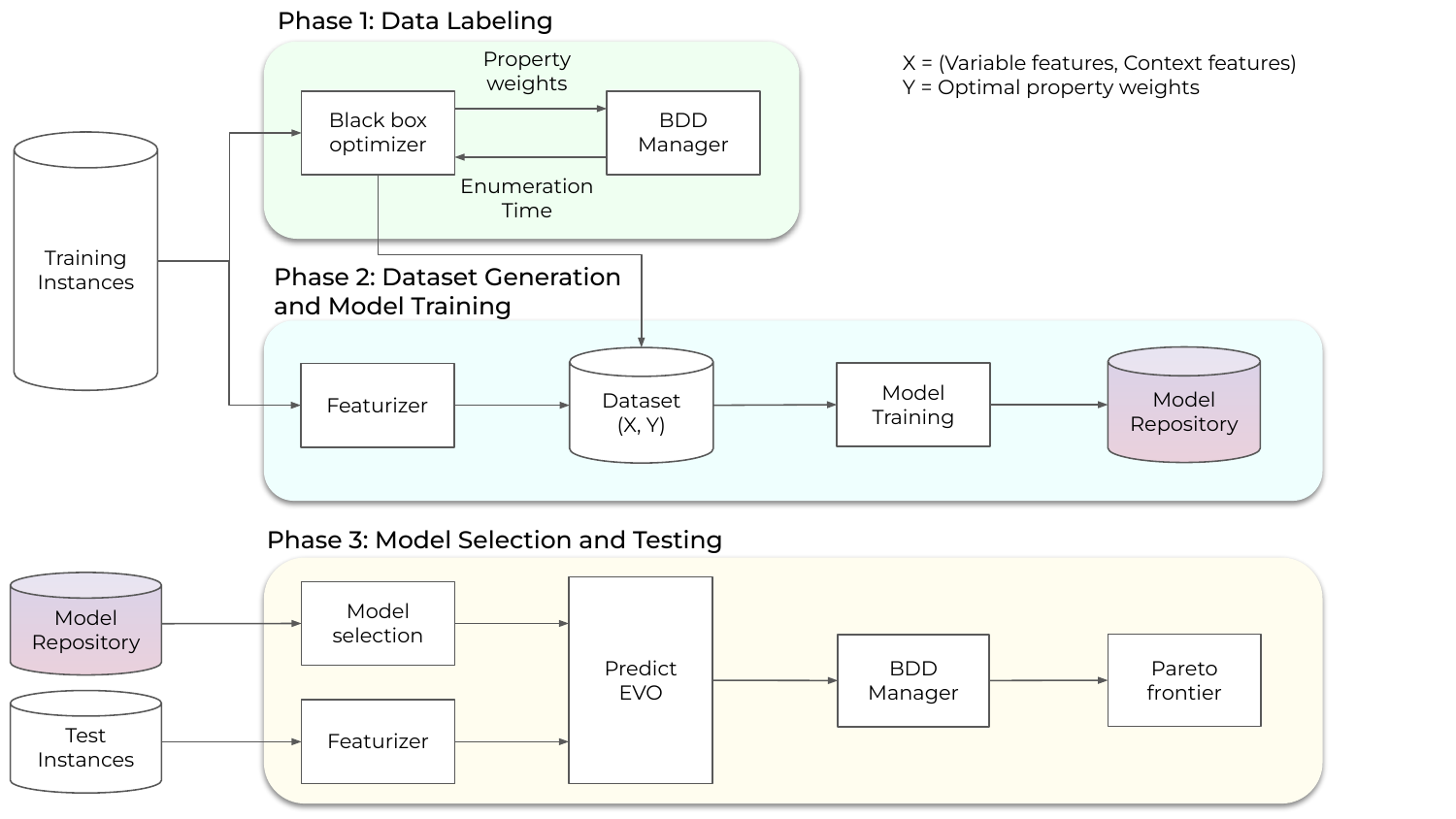}
    \caption{Schematic of the proposed method \method{}, which comprises of three phases. In Phase 1, we generate property-weight labels by iteratively improving them using the interplay between ``Black-box optimizer'' and ``BDD Manager''. Phase 2 focuses on building datasets of tuples of features extracted by ``Featurizer`` and best property weights obtained from Phase 1 for training learning-to-rank models. In Phase 3 we do ``Model selection'' to select the model with best generalization and use it to predict efficient variable order. Finally, we use the predicted variable order to construct the BDD and enumerate the Pareto frontier.}
    \label{fig:block_diagram}
\end{figure*}

\section{Methodology}
\label{sec:method}
The proposed methodology, as depicted in \Cref{fig:block_diagram}, is divided into three phases. 
We apply our technique to the multiobjective knapsack problem (MKP) which can be described as: 
$$\max_{x \in \{0, 1\}^{n}}\left \{ \left \{ \sum_{i \in [n]} a^p_i x_i \right \}_{p=1}^P : \sum_{i \in [n]} w_i x_i \leq W \right \}.$$
Here, $n$ is the number of items/variables, $w_i$ and $a^p_i$ are the weight and profit corresponding to each item $i \in [n]$ and objective $p \in [P]$. Finally, the capacity of the knapsack is $W\in\mathbb{Z}_{+}$.
Next, we define an efficient variable ordering, a concept which will be useful in describing~\method{}.

Let $\mathcal O$ be the set of all possible variable orderings and $\Gamma(o), o \in \mathcal O$ denote the \paretofront{} enumeration time over a BDD constructed using order $o$.
Let $o^\star \equiv \argmin_{o \in \mathcal O} \Gamma(o)$ be the optimal \vo{}.
Finding $o^\star$ among all $n!$ possible permutations of $n$ variables is intractable, so we will aim for an efficient variable ordering (\evo) $o^e$ that is as close as possible to $o^\star$. Note that our approach is heuristic and does not come with approximation guarantees on the enumeration time of $o^e$ relative to $o^\star$.

The objective in the first phase is to find, for each training instance, an EVO that acts as a label for supervising an ML model.
In the second phase, each training instance is mapped to a set of features and an ML model is trained.
Finally, we perform model selection and use the chosen model to predict EVOs, referred to as $\hat o^e$, that are then used to construct a BDD and compute the \paretofront{} for any test instance.

\begin{table}[htbp!]
    \centering
    \begin{tabular}{lcl}
    \toprule
        Property & &Definition  \\
    \midrule
     weight & \qquad~ &$w_i$ \\ [4pt]
     avg-value & \qquad~ & $\sum_{p = 1}^P a^p_i / P$ \\ [4pt]
     max-value & \qquad~ & $\max \{a^p_i\}_{p=1}^P$ \\ [4pt]
     min-value & \qquad~ & $\min \{a^p_i\}_{p=1}^P$ \\[4pt]
     avg-value-by-weight & \qquad~ & $\left(\sum_{p = 1}^P a^p_i / P\right) / w_i$ \\ [4pt]
     max-value-by-weight & \qquad~ & $\max \{a^p_i\}_{p=1}^P / w_i$ \\ [4pt]
     min-value-by-weight & \qquad~ & $\min \{a^p_i\}_{p=1}^P / w_i$ \\
     \bottomrule
    \end{tabular}
    \caption{Properties of a variable $i$ of an MKP.}
    \label{tab:prop_mkp}
\end{table}


\subsection{Phase 1: Finding an EVO}
\label{sec:label_moo}
Since finding an optimal \vo{} that minimizes the \paretofront{} enumeration time is NP-complete, we devise a heuristic approach for finding an \evo{}, $o^e$.

To find $o^e$ for a given instance $\mathcal I$, we use black-box optimization as the run time $\Gamma^{\mathcal I}$ cannot be described analytically and optimized over.
A naive approach to find $o^e$ would be to search directly in the variable ordering space, as suggested in Approach \#1 of \Cref{fig:approaches}. While this might work well for tiny problems, it will not scale with the increase in the problem size as there are $n!$ possible orderings. 

To alleviate this issue, we define a surrogate search space that removes the dependence on the problem size. 
Specifically, we introduce score-based variable ordering where we order the variables based on the decreasing order of their total score. 
For a given problem class, we define a set of properties $\mathcal K$ for its variables that capture some problem structure.
For example, in an MKP, the weight of an item can act as property.
\Cref{tab:prop_mkp} lists all properties of a variable of an MKP.
Let $g_{ik}$ be the property score of variable $i$ for some property $k \in \mathcal K$, $\bar w = (w_1, \cdots, w_k)$ be the property weights in $[-1, 1]^{|\mathcal K|}$.
Then, the score of a variable $i$ is defined as 
\begin{equation}
s_i \equiv \sum_{k \in \mathcal K} w_k \cdot \frac{g_{ik}}{\sum_{i \in [n]}{g_{ik}}}.
\end{equation}
We recover the variable ordering by sorting them in decreasing order of their score.
Thus, as depicted in Approach \#2 in \Cref{fig:approaches}, the search for $o^e$ is conducted in the surrogate search space $[-1, 1]^{|\mathcal K|}$, which only depends on $|\mathcal K|$ and not on the number of variables $n$.
Note that defining the search space in this manner gives an additional layer of dynamism in the sense that two instances with the same property weights can have different variable orders.

With a slight abuse of notation, let $\Gamma(\bar w)$ represent the time taken to enumerate the \paretofront{} using a \vo{} obtained by property weights $\bar w$.
Given a problem instance, the black-box optimizer controls the property weights and the BDD manager computes the \paretofront{} enumeration time based on the \vo{} derived from these property weights. 
The black-box optimizer iteratively tries different property weight configurations, maintaining a list of incumbents, a process that is depicted in Phase 1 of \Cref{fig:block_diagram}.
We propose to use the variable ordering obtained from the best incumbent property weight as the label for learning task.

\begin{table*}[htbp!]
    \centering
    {\small
    \begin{tabular}{llll}
        \toprule
        Type & Feature & \qquad Description & \quad Count \\
        \midrule
        \multirow{2}{*}{Variable} & Properties & \qquad The variable properties described in \Cref{tab:prop_mkp} & \quad 7 \\
        \cmidrule{2-4}
        & Rank & \qquad The ranks corresponding to the heuristic orderings & \quad 10\\
        & & \qquad described in \Cref{tab:ratio_of_sums} & \quad \\
        \cmidrule{2-4}
        & Value SD & \qquad The standard deviation of the values & \quad 1\\
        \cmidrule{1-4}
        \multirow{7}{*}{Context} & \# objectives & \qquad The number of objectives in the MKP problem & \quad 1\\
        \cmidrule{2-4}
        & \# items & \qquad The number of items in the MKP problem & \quad 1\\
        \cmidrule{2-4}
        & Capacity & \qquad The capacity of the MKP & \quad 1\\
        \cmidrule{2-4}
        & Weight stats. & \qquad The mean, min., max. and std. of the MKP weights & \quad 4\\
        \cmidrule{2-4}
        & Aggregate value & \qquad The mean, min., max. and std. of the values & \quad 12 \\
        & stats. & \qquad of each objective & \quad \\
        \midrule
        Total count & & & \quad 37\\
        \bottomrule
    \end{tabular}
    }
    \caption{Features for an MKP.}
    \label{tab:mkp_features}
\end{table*}

\subsection{Phase 2: Dataset Generation and Model Training}
In this phase, we begin by generating the training dataset.
We give special attention to designing our features such that the resulting models are permutation-invariant and independent of the size of the problem. Note that instead of this feature engineering approach, one could use feature learning through graph neural networks or similar deep learning techniques, see~\cite{cappart2021combinatorial} for a survey. However, given that our case study is on the knapsack problem, we opt for domain-specific feature engineering that can directly exploit some problem structure and leads to somewhat interpretable ML models.

Suppose we are given $J$ problem instances, each having $n_j$ variables, $j\in[J]$. 
Let $\alpha_{ij}$ denote the features of variable $i\in[n_j]$ and $\beta_{j}$ denote the instance-level context features.
Using the features and the EVO computed in Phase 1, we construct a dataset 
$\mathcal D= \left\{(\alpha_{ij}, \beta_{j}, r_i({o^e}_{j}): i \in [n_j] : j \in [J]\right\}$.
Here, $o^e_{j}$ is the EVO of an instance $j$ and $r_i: \mathbb Z^{n_j}_+ \rightarrow \mathbb Z_+$ a mapping from the  EVO to the rank of a variable $i$.
For example, if $n_j = 4$ for some instance $j$ and $o^e_{j} = (2, 1, 4, 3)$, then $r_1( o^e_{j})=3$, $r_2( o^e_{j})=4$, $r_3( o^e_{j})=1$, and $r_4( o^e_{j})=2$.
For a complete list of variable and context features, refer to \Cref{tab:mkp_features}. 

\noindent\textbf{Learning-to-rank} (LTR) is an ML approach specifically designed for ranking tasks, where the goal is to sort a set of items based on their relevance given a query or context. It is commonly used in applications such as information retrieval.
We formulate the task of predicting the EVO as an LTR task and use the pointwise and pairwise ranking approaches to solve the problem.

In the pointwise approach, each item (variable) in the training data is treated independently, and the goal is to learn a model that directly predicts the relevance score or label for each variable. This is similar to solving a regression problem with a mean-squared error loss. 
Specifically, we train a model $f_{\theta}(\alpha_{ij}, \beta_j)$ to predict $r_i(o^e_j)$.
Once the model is trained, it can be used to rank items based on their predicted scores $\hat o^e$. However, this approach does not explicitly consider the relationships between variables in its loss function.

The pairwise approach aims to resolve this issue~\cite{joachims2002optimizing}. 
Let $$\mathcal T_j = \left\{(i_1, i_2): r_{i_1}(o^e_j) > r_{i_1}(o^e_j))\right\}$$ be the set of all variable tuples $(i_1, i_2)$ such that $i_1$ is ranked higher than $i_2$ for instance $j$.
Then, the goal is to learn a model that maximizes the number of respected pairwise-ordering constraints. 
Specifically, we train a model $g_{\phi}(\cdot)$ such that the number of pairs $(i_1, i_2) \in \mathcal T_j$ for which $g_{\phi}(\alpha_{i_1j}, \beta_j) > g_{\phi}(\alpha_{i_2j}, \beta_j)$ is maximized. 
This approach is better equipped to solve the ranking problem as the structured loss takes into account the pairwise relationships.


\subsection{Phase 3: Model Selection and Testing}
We follow a two-step approach to perform model selection as the ranking task is a proxy to the downstream task of efficiently solving a multiobjective problem.
Firstly, for each model class (e.g., decision trees, linear models, etc.), we select the best model based on Kendall's Tau~\cite{kendall1938new}, a ranking performance metric that measures the fraction of violated pairwise-ordering constraints, on instances from a validation set different from the training set. 
Subsequently, we pit the best models from each type against one another and select the winner based on the minimum average \paretofront{} enumeration time on the validation set. 
Henceforth, for previously unseen instances from the test set, we will use the model selected in Phase 3 to predict the  EVO and then compute the \paretofront{}.

%% file: Sections/study.tex
\section{Computational Setup}
\label{sec:setup}
Our code and instances are available at~\url{https://github.com/khalil-research/leo}.
All the experiments reported in this manuscript are conducted on a computing cluster with an Intel Xeon CPU E5-2683 CPUs.
We use \smac{} \citep{lindauer2021smac3} -- a black-box optimization and algorithm configuration library -- for generating the labels. The ML models are built using Python 3.8, Scikit-learn \cite{scikit-learn}, and XGBoost \cite{Chen:2016:XST:2939672.2939785}, and SVMRank \cite{joachims2006training}.
The ``BDD Manager'' is based on the implementation of 
\citet{bergman2016multiobjective}, which is available at \url{https://www.andrew.cmu.edu/user/vanhoeve/mdd/}.

\subsection{Instance Generation}
We use a dataset of randomly generated MKP instances as described in \cite{bergman2016multiobjective}.
The values $w_i$ and $a^p_i$ are sampled randomly from a discrete uniform distribution ranging from 1 to 100. 
The capacity $W$ is set to $\left\lceil 0.5 \sum_{i \in I} w_i \right \rceil$.
We generate instances with sizes $$\mathcal S = \{(3, 60), (3, 70), (3, 80), (4, 50), (5, 40), (6, 40), (7, 40)\},$$ where the first and second component of the tuple specify the number of objectives and variables, respectively.
For each size, we generate 1000 training instances, 100 validation instances, and 100 test instances.

\subsection{Instrumenting SMAC}
\noindent\textbf{As a labeling tool:} 
To generate EVOs for the learning-based models, we use \smac{}~\cite{lindauer2021smac3}. In what follows, \smacI{} refers to the use of SMAC as a black-box optimizer that finds an EVO for a given training instance.
Specifically, \smacI{} solves $\bar w^e_j = \min_{\bar w} \Gamma_j(\bar w)$ for each instance $j$ in the training set; we obtain $o^e_j$ by calculating variable scores -- a dot product between $\bar w^e_j$ and the corresponding property value -- and sorting variables in the decreasing order of their score. 

\noindent\textbf{As a baseline:} 
The other more standard use of SMAC, which we refer to as~\smacD{}, is as an algorithm configuration tool. To obtain a \smacD{} ordering, we use \smac{} to solve $\bar w^{e}_D = \min_{\bar w} \mathbb E_{\bm{j \sim [J]}}\left[\Gamma_j(\bar w)\right]$ and obtain an order $o^e_{D_j}$ for instance $j$ using single property weight vector $\bar w^{e}_D$. The expectation of the run time here simply represents its average over the $|J|$ training instances.
Note that we get only one property weight vector for the entire dataset in the \smacD{} case rather than one per instance as in the \smacI{} case.
However, we obtain an instance-specific VO when using the \smacD{} configuration as the underlying property values change across instances.

\noindent\textbf{Initialization:} 
For both uses of SMAC, we use the \minWt{} ordering as a warm-start by assigning all the property weights to zero except the weight property, which is set to -1. This reduces the need for extensive random exploration of the configuration space by providing a reasonably good ordering heuristic. 

\noindent\textbf{Random seeds:} 
As SMAC is a randomized optimization algorithm, running it with multiple random seeds increases the odds of finding a good solution. We leverage this idea for \smacI{} as its outputs will be used as labels for supervised learning and are thus expected to be of very high quality, i.e., we seek instance-specific parameter configurations that yield variable orderings with minimal solution times. We run \smacD{} with a single seed and use its average run time on the training set as a target to beat for \smacI{}. Since the latter optimizes at the instance level, one would hope it can do better than the distribution-level configuration of \smacD{}.

As such, for \smacI{}, we run \smac{} on each instance with 5 seeds for all sizes except (5, 40) and (6, 40), for which we used one seed. 
We start with one seed, then average the run time of the best-performing seed per instance to that of the average enumeration time of \smacD{} on the training set, and relaunch \smac{} with a new seed until a desired performance gap between \smacI{} and \smacD{} is achieved. 

\noindent\textbf{Computational budget:} 
In the \smacD{} setting, we run \smac{} with a 12-hour time limit, whereas in the \smacI{} case the time limit is set to 20 minutes per instance except for sizes (3, 60), (4, 50), (5, 40), for which it is set to 5 minutes per instance. 
In both settings, \smac{} runs on a 4-core machine with a 20GB memory limit.
It can be observed that generating the labels can be computationally expensive. 
This dictates the choice of sizes in the set of instance sizes, $\mathcal S$.
Specifically, we select instance sets with an average running time of at most 100 seconds (not too hard) and at least 5 seconds (nor too easy) using the top-down compilation method described in \citet{bergman2021network}.

\subsection{Learning Models}
We use linear regression, ridge regression, lasso regression, decision trees, and gradient-boosted trees (GBT) with mean-squared error loss to build size-specific pointwise ranking models.
Similarly, we train support vector machines and GBT with pairwise-ranking loss to obtain pairwise ranking models for each size. 
In the experimental results that follow, the best learning-based method will be referred to as \ml. This turns out to be GBT trained with pairwise-ranking loss. 
The GBT models that were selected achieved a Kendall's Tau ranging between 0.67 to 0.81 across all problem sizes on the validation set. 
The model selection follows the procedure mentioned in Phase 3 of the Methodology section. 

In terms of features, we omit the context features in~\Cref{tab:mkp_features} when training linear size-specific models, as these features take on the same value for all variables of the same instance and thus do not contribute to the prediction of the rank of a variable. The context features are used with non-linear models such as decision trees and GBT.

We also train two additional GBT models -- ML+A and ML+AC -- with pairwise-ranking loss, on the \textit{union of the datasets of all sizes}. 
In particular, ML+A is trained with only variable features, similar to~\ml{}, whereas ML+AC adds the instance context features to the variable features.

\subsection{Baselines}
To evaluate the performance of learning-based orderings, we compare to four baselines:
\begin{itemize}
    \item[--] \textbf{\lex{}} uses the (arbitrary) default variable ordering in which the instance was generated. 

    \item[--] \textbf{\minWt{}} orders the variables in increasing order of their weight values, $w_i$. This is a commonly used heuristic for solving the single-objective knapsack problem. 

    \item[--] \textbf{\minValByWt{}} orders the variables in decreasing order of the property min-value-by-weight detailed in \Cref{tab:prop_mkp}, which is defined as  $\min \{a^p_i\}_{p=1}^P / w_i$. This rule has an intuitive interpretation: it prefers variables with larger \textit{worst-case} (the minimum in the numerator) value-to-weight ratio. It is not surprising that this heuristic might perform well given that it is part of a $\frac{1}{2}$-approximation algorithm for the single-objective knapsack problem~\cite{chekuri-notes}.

    \item[--] \textbf{\smacD{}}, as described in an earlier paragraph. It produces one weight setting for the property scores per instance set. This baseline can be seen as a representative for the algorithm configuration paradigm recently surveyed by~\citet{schede2022survey}.

\end{itemize}

%% file: Sections/results.tex
\begin{table*}[htbp!]
    \centering
    \resizebox{\textwidth}{!}{%
    \begin{tabular}{r l l l l l l l l l l}
        \toprule
        Heuristic ordering & \multicolumn{10}{c}{Size}\\
        \cmidrule{2-11}
        \textlangle sort\textrangle\_\textlangle property-name\textrangle & ~~(3, 20) & (3, 40) & (3, 60) & (3, 80) & (5, 20) & (5, 30) & (5, 40) & (7, 20) & (7, 30) & (7, 40)\\
        \midrule
        max\_weight & ~~\textbf{0.778} & 1.147 & 1.378 & 1.317 & 0.815 & 1.865 & 2.760 & 1.165 & 2.769 & 1.830 \\ 
        min\_weight & ~~0.812 & \textbf{0.654} & 0.648 & 0.685 & \textbf{0.769} & \textbf{0.700} & \textbf{0.533} & \textbf{0.694} & \textbf{0.588} & \textbf{0.581} \\
        max\_avg-value & ~~0.861 & 0.891 & 0.843 & 0.972 & 0.839 & 0.992 & 0.840 & 0.845 & 1.125 & 0.823 \\   
        min\_avg-value & ~~0.893 & 1.619 & 2.144 & 1.692 & 0.936 & 1.721 & 2.132 & 1.103 & 1.918 & 1.524 \\   
        max\_max-value & ~~0.858 & 1.065 & 1.126 & 1.203 & 0.857 & 1.137 & 1.060 & 0.899 & 1.408 & 0.992 \\   
        min\_max-value & ~~0.855 & 1.018 & 1.010 & 0.995 & 0.854 & 1.077 & 1.009 & 0.909 & 0.981 & 0.975 \\   
        max\_min-value & ~~0.851 & 0.705 & \textbf{0.537} & \textbf{0.590} &0.813 & 0.816 & 0.617 &  0.819 & 0.862 & 0.767 \\ 
        min\_min-value & ~~0.883 & 1.662 & 2.266 & 1.729 &0.930 & 1.700 & 2.020 & 1.094 & 1.980 & 1.430\\  
        max\_avg-value-by-weight & ~~0.838 & 0.832 & 0.988 & 1.156 &0.802 & 0.896 & 0.887 &  0.751 & 0.917 & 0.788 \\   
        max\_max-value-by-weight & ~~0.837 & 0.837 & 1.037 & 1.192 &0.804 & 0.871 & 0.824 & 0.732 & 0.790 & 0.735 \\       
        \bottomrule
    \end{tabular}    
    }
    \caption{Ratio of the average time taken by a heuristic ordering to that of 5 random orderings across 250 MKP instances across sizes specified. A ratio of less (greater) than 1 indicates that heuristic ordering performs better (worse) than random ordering, on average. The name of a heuristic ordering follows the structure \textlangle sort\textrangle\_\textlangle property-name\textrangle, where \textlangle sort\textrangle~can either be min or max and \textlangle property-name\textrangle~can be equal to one of the properties defined in \Cref{tab:prop_mkp}. A value of max (min) for \textlangle sort\textrangle~would sort the variables in the descending (ascending) order of the property \textlangle property-name\textrangle.}  
    \label{tab:ratio_of_sums}
\end{table*}

\section{Experimental Results}
\label{sec:results}

We examine our experimental findings through a series of questions that span the impact of \vo{} on PF enumeration time (Q1), the performance of~\smacI{} as a black-box optimizer that provides labels for ML (Q2), the performance of~\method{} on unseen test instances and comparison to the baselines (Q3, Q4), a feature importance analysis of the best ML models used by~\method{} (Q5), and an exploration of size-independent, unified ML models (Q6).

\subsubsection{Q1. Does variable ordering impact the Pareto frontier enumeration time?} 

To test the hypothesis that VO has an impact on the PF enumeration time, we compare the run time of ten heuristic orderings against the expected run time of random orderings. By the run time of an ordering, we mean the time taken to compute the PF on a BDD constructed using that particular ordering. 

The heuristic orderings are constructed using the variable properties. For example, the heuristic ordering called max\_avg-value-by-weight sorts the variables in descending order of the ratio of a variable's average value by its weight. We estimate the expected run time of a random ordering by sampling 5 heuristic variable orderings uniformly at random from all possible $n!$ orderings and averaging their run time.
\Cref{tab:ratio_of_sums} summarizes the results of this experiment. We work with 250 MKP instances of the problem sizes (3, 20), (3, 40), (3, 60), (3, 80), (5, 20), (5, 30), (5, 40), (7, 20), (7, 30), and (7, 40). The values in the table are the ratios of the average run time of a heuristic ordering to that of the 5 random orderings; values smaller than one indicate that a heuristic ordering is faster than a random one, on average.
The best heuristic ordering for each size is highlighted.


{First, it is clear that some heuristic orderings consistently outperform the random ones across all problem sizes (min\_weight, max\_min-value), and by a significant margin. In contrast, some heuristic orderings are consistently worse than random.
For example, the heuristic ordering max\_weight, min\_avg-value, and min\_min-value consistently underperform when the number of variables is more than 20}.

{Second, the choice of \minWt{} as a baseline was motivated by the results of this experiment as  min\_weight wins on most sizes as highlighted in \Cref{tab:ratio_of_sums}}.

Altogether, this experiment validates the hypothesis that \vo{} has an impact on the PF enumeration time and that there exists \evo{} that can significantly reduce this time.




\begin{table}[htbp!]
    \centering
    \resizebox{0.95\linewidth}{!}{    
    \begin{tabular}{llrrrr}
    \toprule
       Size &   Method &  Count &  GMean &   Min &     Max \\
    \midrule
    \midrule
    \multirow{5}{*}{(5, 40) }&       ML &    100 &   1.43 &  1.26 &   40.91 \\
    &    ML+AC &    100 &   1.54 &  1.06 &   41.29 \\
    &     \textbf{ML+A} &    100 &   \textbf{1.38} &  1.28 &   41.24 \\
    &    SmacD &    100 &   2.49 &  1.38 &   51.22 \\
    & MaxRatio &    100 &   2.48 &  1.44 &   58.87 \\
    &    MinWt &    100 &   2.42 &  1.44 &   52.61 \\
    &      Lex &    100 &   6.23 &  1.86 &  109.18 \\
    \midrule
    \multirow{5}{*}{(6, 40) }&       ML &    100 &  14.45 &  2.57 &  377.89 \\
    &    \textbf{ML+AC} &    100 &  \textbf{13.62} &  2.01 &  427.80 \\
    &     ML+A &    100 &  13.67 &  2.29 &  355.64 \\
    &    SmacD &    100 &  18.62 &  2.95 &  434.34 \\
    & MaxRatio &    100 &  19.84 &  3.05 &  664.88 \\
    &    MinWt &    100 &  18.82 &  2.91 &  427.06 \\
    &      Lex &    100 &  42.23 &  6.01 &  713.58 \\
    \midrule
    \multirow{5}{*}{(7, 40) }&       \textbf{ML} &    100 &  \textbf{41.97} &  2.26 & 1490.58 \\
    &    ML+AC &     99 &  45.39 &  2.27 & 1153.07 \\
    &     ML+A &     99 &  44.70 &  2.81 & 1130.82 \\
    &    SmacD &     99 &  64.84 &  2.95 & 1521.43 \\
    & MaxRatio &     98 &  86.49 &  4.14 & 1700.44 \\
    &    MinWt &     99 &  60.08 &  3.16 & 1561.36 \\
    &      Lex &     95 & 102.23 &  7.66 & 1604.17 \\
    \midrule
    \multirow{5}{*}{(4, 50) }&       \textbf{ML} &    100 &   \textbf{5.75} &  2.54 &   79.42 \\
    &    ML+AC &    100 &   5.97 &  2.22 &   77.75 \\
    &     ML+A &    100 &   6.04 &  2.47 &   73.17 \\
    &    SmacD &    100 &  12.17 &  4.04 &  218.24 \\
    & MaxRatio &    100 &  14.23 &  3.24 &  167.95 \\
    &    MinWt &    100 &  18.37 &  4.67 &  553.15 \\
    &      Lex &    100 &  32.25 &  6.67 &  395.67 \\
    \midrule
    \multirow{5}{*}{(3, 60) }&       ML &    100 &   3.15 &  3.88 &   34.13 \\
    &    \textbf{ML+AC} &    100 &   \textbf{3.07} &  3.44 &   31.99 \\
    &     ML+A &    100 &   3.23 &  3.56 &   31.14 \\
    &    SmacD &    100 &   4.07 &  3.53 &   43.58 \\
    & MaxRatio &    100 &   9.24 &  5.76 &   64.21 \\
    &    MinWt &    100 &  11.61 &  5.36 &   75.98 \\
    &      Lex &    100 &  20.19 &  8.12 &  195.25 \\
    \midrule
    \multirow{5}{*}{(3, 70) }&       ML &    100 &  15.05 &  7.92 &   64.55 \\
    &    \textbf{ML+AC} &    100 &  \textbf{14.94} &  7.41 &   65.67 \\
    &     ML+A &    100 &  15.35 &  7.68 &   80.82 \\
    &    SmacD &    100 &  16.64 &  7.52 &   96.94 \\
    & MaxRatio &    100 &  33.54 & 12.01 &  215.63 \\
    &    MinWt &    100 &  44.43 & 12.17 &  194.07 \\
    &      Lex &    100 &  71.55 & 19.87 &  308.21 \\
    \midrule
    \multirow{5}{*}{(3, 80) }&       ML &    100 &  43.85 & 15.29 &  286.11 \\
    &    \textbf{ML+AC} &    100 &  \textbf{42.89} & 15.43 &  262.57 \\
    &     ML+A &    100 &  44.64 & 15.35 &  262.77 \\
    &    SmacD &    100 &  53.33 & 15.66 &  632.92 \\
    & MaxRatio &    100 & 100.31 & 29.64 &  741.82 \\
    &    MinWt &    100 & 129.40 & 30.35 &  794.10 \\
    &      Lex &     99 & 196.31 & 56.52 & 1332.19 \\
    \bottomrule
    \end{tabular}}
    \caption{\paretofront{} enumeration time evaluation results on the test set with a time limit of 1800 seconds. 
    Each row represents aggregate statistics for a given instance ``Size'' and ``Method''. 
    ``Count'' stands for the number of instances on which the algorithm ran successfully without hitting the time or memory limits. ``GMean'', ``Min'' and ``Max'' denotes the geometric mean, minimum and maximum of the enumeration time computed across ``count'' many instances. We use a shift of 5 to compute the ``GMean''.}
    \label{tab:knapsack_test_time}
\end{table}

\subsubsection{Q2. Can~\smacI{} find good variable orderings for training instances?}
\begin{figure*}[htbp!]
    \centering
    \includegraphics[width=\linewidth]{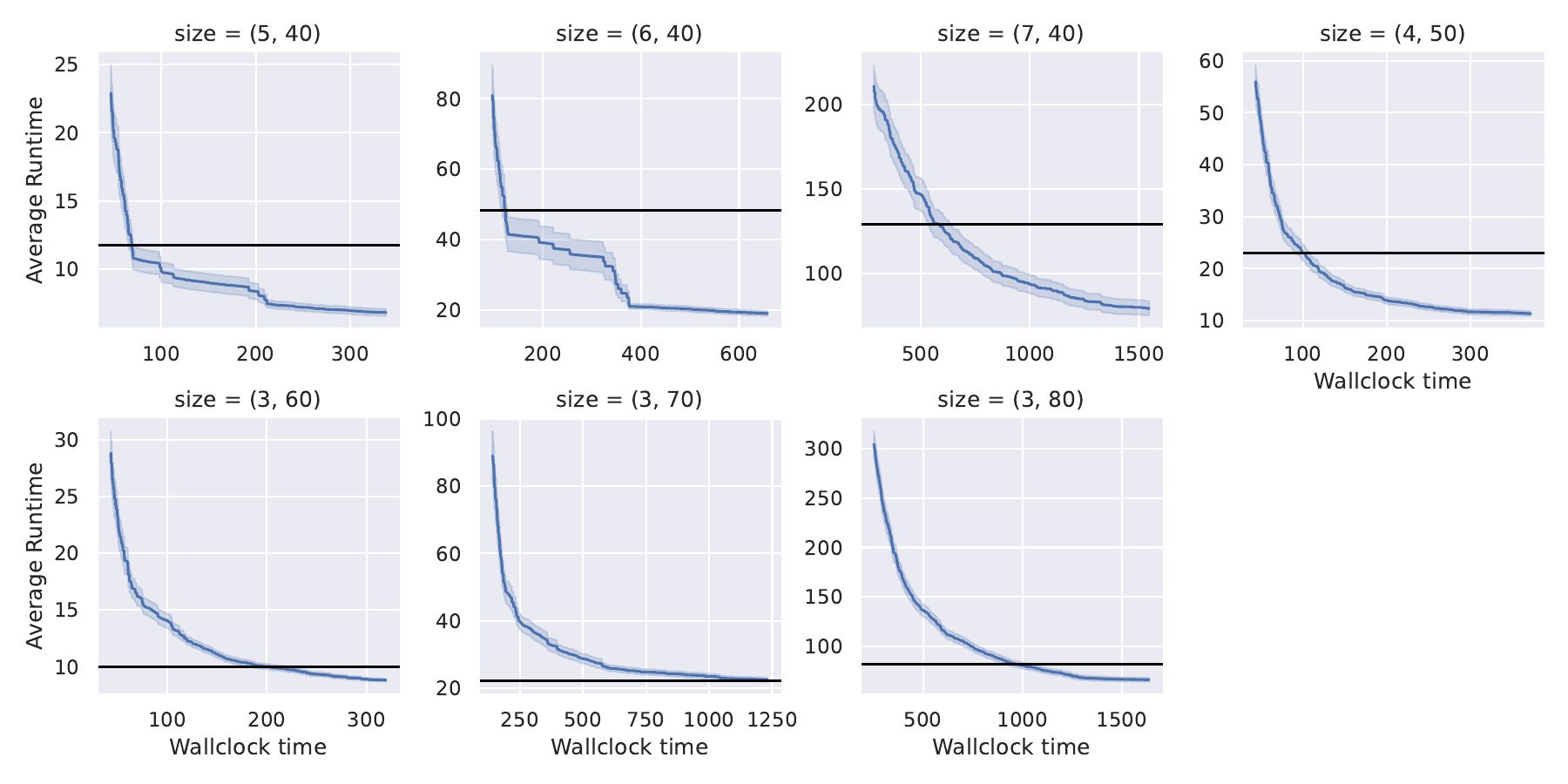}
    \caption{SmacI performance w.r.t. wallclock time across different sizes. For a given size, we first of all find the best seed, i.e., the seed which finds an incumbent property-weight having minimum \paretofront{} enumeration time, for each instance. Then for a given wallclock time, we calculate the mean and standard error of the \paretofront{} enumeration time across instances for the incumbent property-weight up to that wallclock time. 
    The mean and standard error are plotted as a blue line and blue error band around it, respectively. The horizontal black line represents the average \paretofront{} enumeration time of \smacD{} configuration on the validation set.}
    \label{fig:smacI_performance}
\end{figure*}
Having established in Q1 that the search for high-quality variable orderings is justified, we now turn to a key component of Phase 1: the use of~\smac{} as a black-box optimizer that produces ``label'' variable orderings for subsequent ML in Phase 2. Figure~\ref{fig:smacI_performance} shows the average run time, on the training instances, of the incumbent property-weight configurations found by~\smac{}. These should be interpreted as standard optimization convergence curves where we seek small values (vertical axis) as quickly as possible (horizontal axis). Indeed,~\smac{} performs well, substantially improving over its initial warm-start solution (the~\minWt{} ordering). The flat horizontal lines in the figure show the average run time of the single configuration found by~\smacD{}. The latter is ultimately outperformed by the~\smacI{} configurations on average, as desired.

\subsubsection{Q3. How does \method{} perform in comparison to baseline methods?}
\Cref{tab:knapsack_test_time} presents the \paretofront{} enumeration across different problem sizes and methods on the test set.
In this question, we will examine only the first of three ML models, namely the first one referred to as~\ml{} in the table; the two other models will be explored in Q7.
We can observe that \ml{} consistently outperforms all baselines in terms of the geometric mean (GMean) of the \paretofront{} enumeration time, across all problem sizes. 

\smacD{} acts as a strong baseline for the \ml{} method, consistently being the second-best in terms of average GMean, except for size (7, 40).
The methods \minWt{} and \minValByWt{} have a nice connection to the single-objective knapsack problem~\cite{chekuri-notes}, as discussed earlier; this might explain why these heuristics reduce the number of intermediate solutions being generated in the BDDs during \paretofront{} enumeration, helping the algorithm terminate quickly.
They closely follow \smacD{} in terms of GMean metric with sizes having more than 3 objectives; however, they are almost twice as worst as \smacD{} on instances with 3 objectives. 

Interestingly, when the number of objectives is larger than 3, \minWt{} outperforms \minValByWt{} (and vice versa). 
This underscores the relationship between the number of objectives and heuristics, i.e., one heuristic might be preferred over another depending on the structure of the problem. 
This also highlights why the \ml{} method performs better than the heuristics as the feature engineering helps create an ensemble of them rather than using only one of them.
Lastly, we observe that method \lex{} has the worst performance across all sizes.

\begin{figure*}[htbp!]
\includegraphics[width=\linewidth]{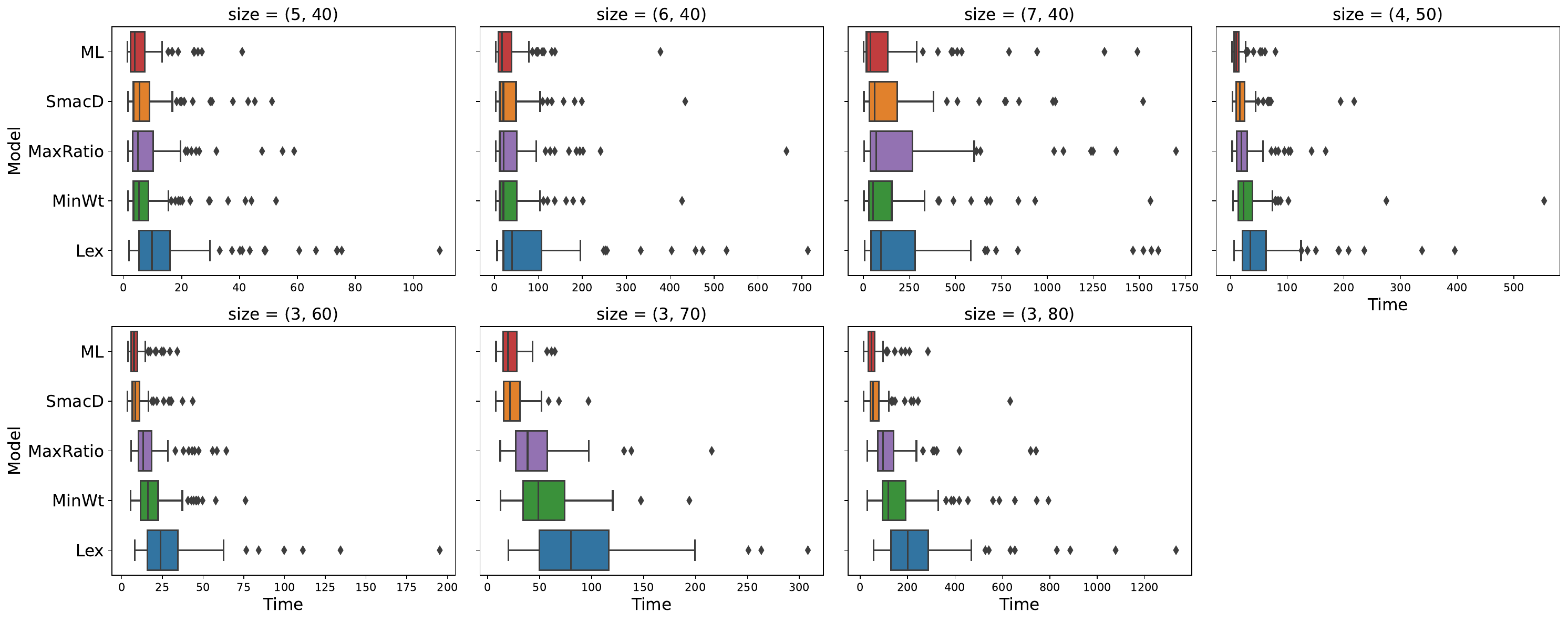} 
\caption{Box plots of time to enumerate the \paretofront{} for different sizes.}
\label{fig:boxplot_size_test}
\end{figure*}



\begin{figure*}[htbp!]
    \centering
    \includegraphics[width=\linewidth]{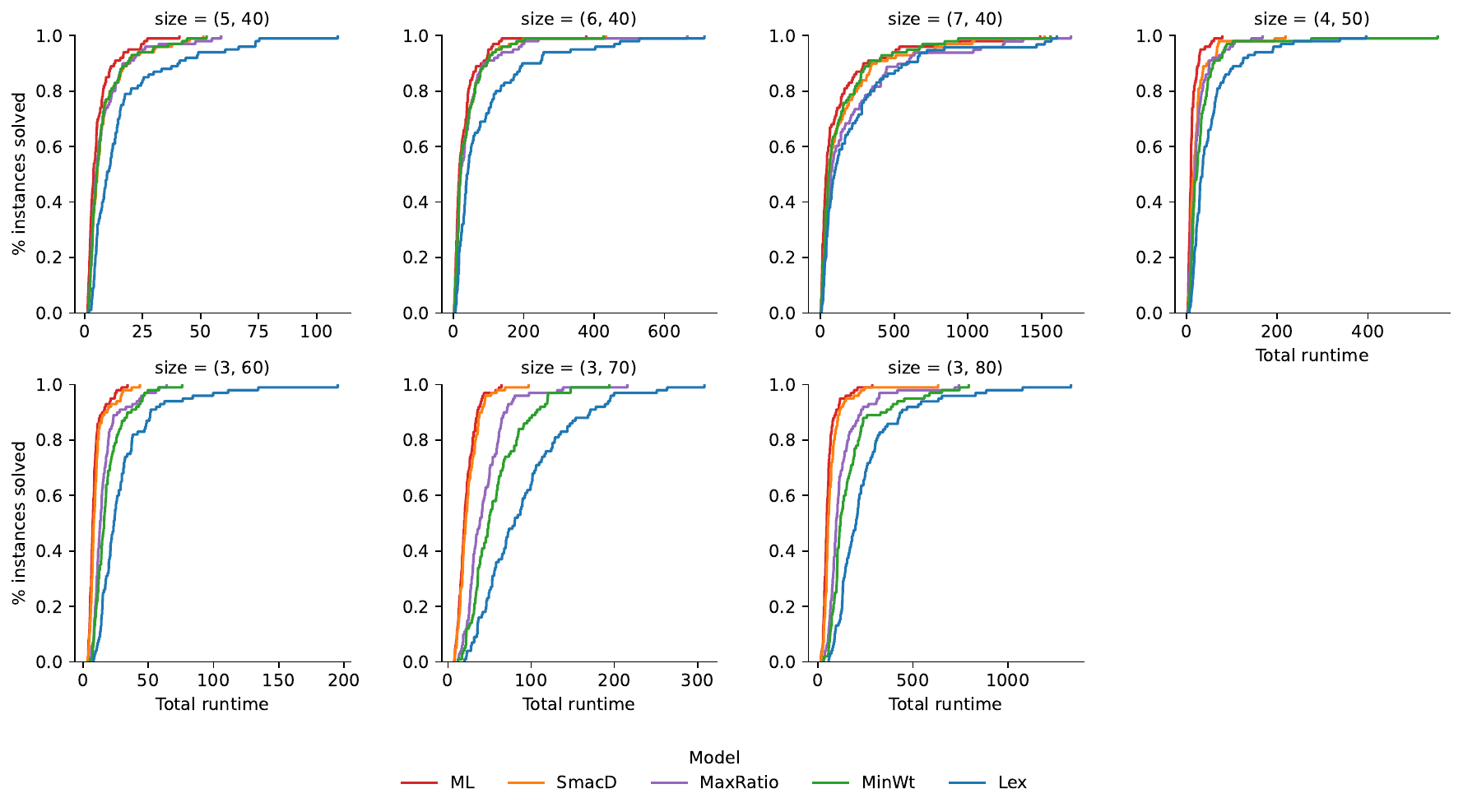}
    \caption{Performance profile of different methods in terms of the fraction of instances solved w.r.t. time for various sizes.}
    \label{fig:perf_size_test}
\end{figure*}

To complement~\Cref{tab:knapsack_test_time}, we present box plots and performance profiles in \Cref{fig:boxplot_size_test} and \Cref{fig:perf_size_test}, respectively.
Note how the \paretofront{} enumeration time distribution for the \ml{} method is much more concentrated, with a smaller median and only a few  outliers compared to other methods. Note that the instances that had a timeout are omitted from this analysis.
This performance improvement leads to a larger fraction of instances being solved in a smaller amount of time as highlighted in \Cref{fig:perf_size_test}.


\subsubsection{Q4. What explains \method{}'s performance?}


\begin{table}[htbp!]
    \centering        
    \resizebox{\linewidth}{!}{
    \begin{tabular}{llrrrr}
    \toprule
    Size &   Method &  Nodes(\%) &  Width(\%) &   Checks(\%) &  GMean(\%) \\
    \midrule
    \midrule
    \multirow{5}{*}{(5, 40) }&       ML &  81.46 &  88.13 &  \textbf{36.28} &  \textbf{22.91} \\
    &    SmacD &  \textbf{69.49} &  \textbf{79.11} &  56.15 &  39.95 \\
    & MaxRatio &  88.80 &  94.10 &  52.49 &  39.75 \\
    &    MinWt &  \textbf{69.49} &  \textbf{79.11} &  56.15 &  38.76 \\
    \midrule
    \multirow{5}{*}{(6, 40) }&       ML &  78.35 &  85.82 &  \textbf{39.36} &  \textbf{34.22} \\
    &    SmacD &  \textbf{70.12} &  \textbf{79.53} &  52.02 &  44.09 \\
    & MaxRatio &  90.63 &  95.01 &  53.39 &  46.97 \\
    &    MinWt &  \textbf{70.12} &  \textbf{79.53} &  52.02 &  44.56 \\
    \midrule
    \multirow{5}{*}{(7, 40) }&       ML &  81.21 &  88.70 &  \textbf{56.99} &  \textbf{41.06} \\
    &    SmacD &  71.18 &  80.18 &  79.12 &  63.43 \\
    & MaxRatio &  72.11 &  81.37 & 100.93 &  84.61 \\
    &    MinWt &  \textbf{70.11} &  \textbf{79.52} &  69.13 &  58.77 \\
    \midrule
    \multirow{5}{*}{(4, 50) }&       ML &  90.60 &  96.68 &  \textbf{21.63} &  \textbf{17.84} \\
    &    SmacD &  76.03 &  87.94 &  42.93 &  37.73 \\
    & MaxRatio &  88.33 &  96.27 &  48.97 &  44.13 \\
    &    MinWt &  \textbf{70.45} &  \textbf{84.91} &  68.81 &  56.97 \\
    \midrule
    \multirow{5}{*}{(3, 60) }&       ML &  93.11 &  98.36 &  \textbf{20.18} &  \textbf{15.58} \\
    &    SmacD &  91.70 &  98.81 &  26.14 &  20.14 \\
    & MaxRatio &  86.32 &  96.42 &  48.45 &  45.76 \\
    &    MinWt &  \textbf{68.68} &  \textbf{86.68} &  62.06 &  57.51 \\
    \midrule
    \multirow{5}{*}{(3, 70) }&       ML &  92.30 &  98.57 &  \textbf{18.21} &  \textbf{21.03} \\
    &    SmacD &  90.81 &  98.71 &  21.52 &  23.26 \\
    & MaxRatio &  86.20 &  97.16 &  49.04 &  46.88 \\
    &    MinWt &  \textbf{70.33} &  \textbf{89.75} &  68.03 &  62.10 \\
    \midrule
    \multirow{5}{*}{(3, 80) }&       ML &  91.09 &  98.69 &  \textbf{19.94} &  \textbf{22.34} \\
    &    SmacD &  92.34 &  99.27 &  27.77 &  27.17 \\
    & MaxRatio &  84.45 &  97.34 &  57.17 &  51.10 \\
    &    MinWt &  \textbf{68.79} &  \textbf{90.58} &  74.72 &  65.91 \\
    \bottomrule
    \end{tabular}}
    \caption{Analyzing the impact of BDD topology on the \paretofront{} enumeration time on the test set.
    For a given ``Size'' and ``Method'', the metric value indicates  the percentage of mean performance by that ``Method'' compared to the \lex{} method. 
    The ``Nodes'' (``Width'') represents the percentage of the mean of the number of nodes (mean width) of the BDDs generated by ``Method'' to that of \lex{}. 
    Similarly, ``Checks'' and ``Time'' indicates the percentage of the average Pareto-dominance checks and the average geometric mean of the \paretofront{} enumeration time, respectively.
    For all metrics, values less than 100 are preferred. For example, a metric with a value of $\delta$\% indicates the ``Method'' is $(100-\delta)$\% better than \lex{}.}
    \label{tab:bdd_topology_analysis}
\end{table}

\begin{figure*}[htbp!]
    \centering
    \includegraphics[width=\linewidth]{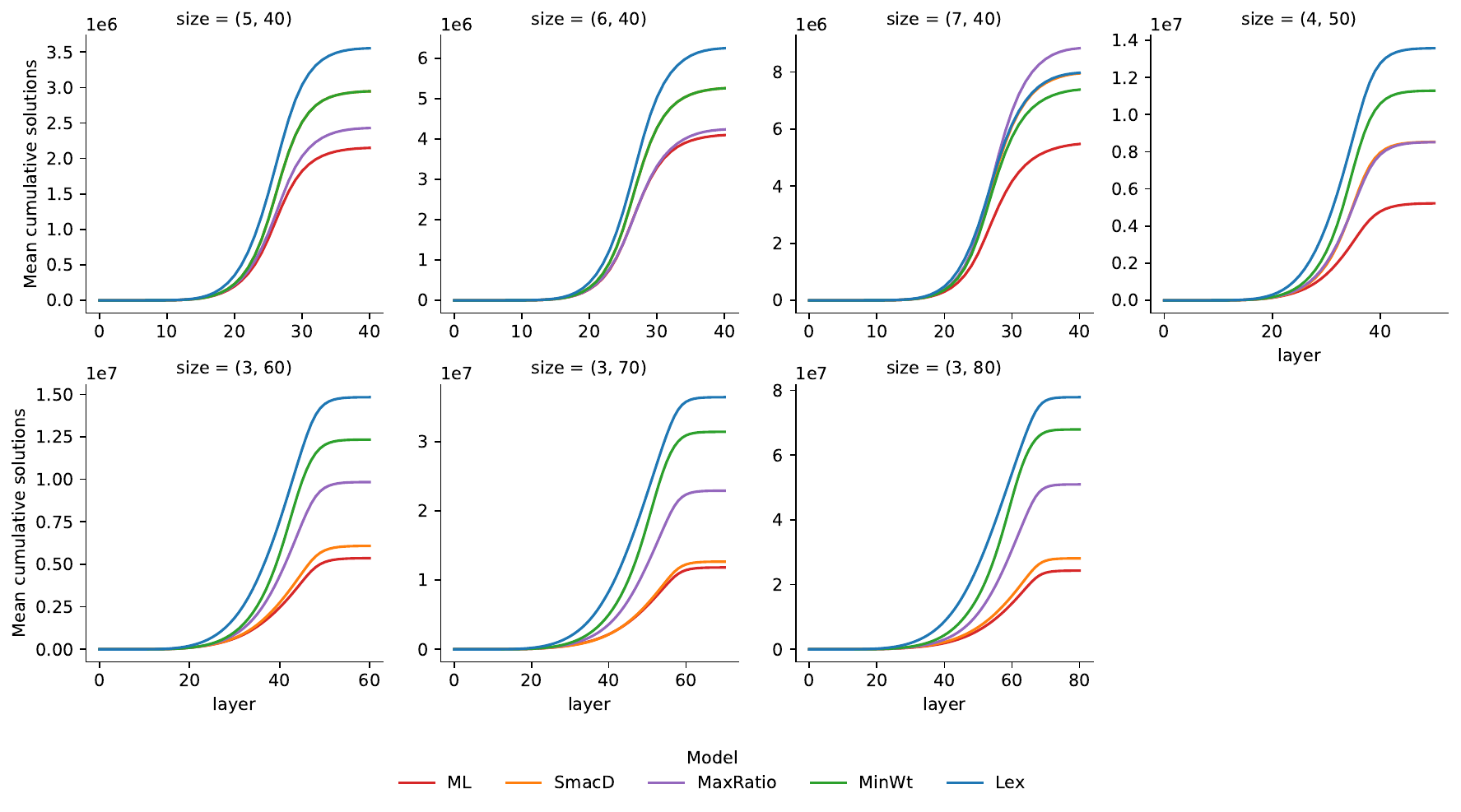}
    \caption{Cumulative number of intermediate solutions generated at a particular layer across all sizes.}
    \label{fig:nnds_profile}
\end{figure*}

Traditionally, smaller-sized exact BDDs are sought after for efficiently performing tasks such as model checking or computing objective function bounds. 
Hoping to find a similar connection in the multiobjective setting, we analyze the relationship between the topology of BDDs generated by different methods and the time to compute the \paretofront{} in \Cref{tab:bdd_topology_analysis}.
Considering the fact that \lex{} performs the worst among all methods and the impact of BDD size on the downstream task, we expect that the size of the BDDs generated by \lex{} to be bigger. 
This holds true across different sizes and methods as it can be observed in \Cref{tab:bdd_topology_analysis} that Nodes and Width values are less than 100.

Extrapolating, one would expect BDDs generated using \ml{} orderings to be the smallest, as it achieves the best performance in terms of enumeration time. 
Counter-intuitively, that is not the case: \minWt{} generates the smallest-sized BDDs on average.
For instance, the value of Nodes for size (3, 80) is 68.79\% for \minWt{}, which is lower than 91.09\% for \ml{}.
However, the reduction in size does not translate to improvements in the running time, as we already know that \ml{} performs best in terms of time. 

We can decipher the performance gains of \ml{} by studying the ``Checks'' metric. 
This metric can be thought of as a proxy of the work done by the \paretofront{} enumeration algorithm; indeed, this metric is positively correlated with Time.
This phenomenon can be further studied in \Cref{fig:nnds_profile}, which shows the mean cumulative solutions generated up to a given layer in the BDD. 
Clearly, \ml{} has the least number of intermediate solutions generated, which also translates to fewer Pareto-dominance checks and smaller Checks. 

To summarize, smaller-sized BDDs can be efficient in enumerating the \paretofront{}. However, reducing the BDD size beyond a certain threshold would inversely affect its performance as it leads to more intermediate solutions being generated, increasing the number of Pareto-dominance checks.
This also validates the need for task-specific methods like \method{} that are specifically geared towards reducing the run time rather than optimizing a topological metric of the BDD.

\subsubsection{Q5. How interpretable are the decisions made by \method{}?}
\method{} uses GBT with the pairwise ranking loss for learning to order variables.
To obtain feature importance scores, we count the number of times a particular feature is used to split a node in the decision trees. 
We then normalize these scores by dividing them by the maximum score, resulting in values in $[0, 1]$ for each size. 

\begin{figure}[htbp!]
    \centering
    \includegraphics[width=\linewidth]{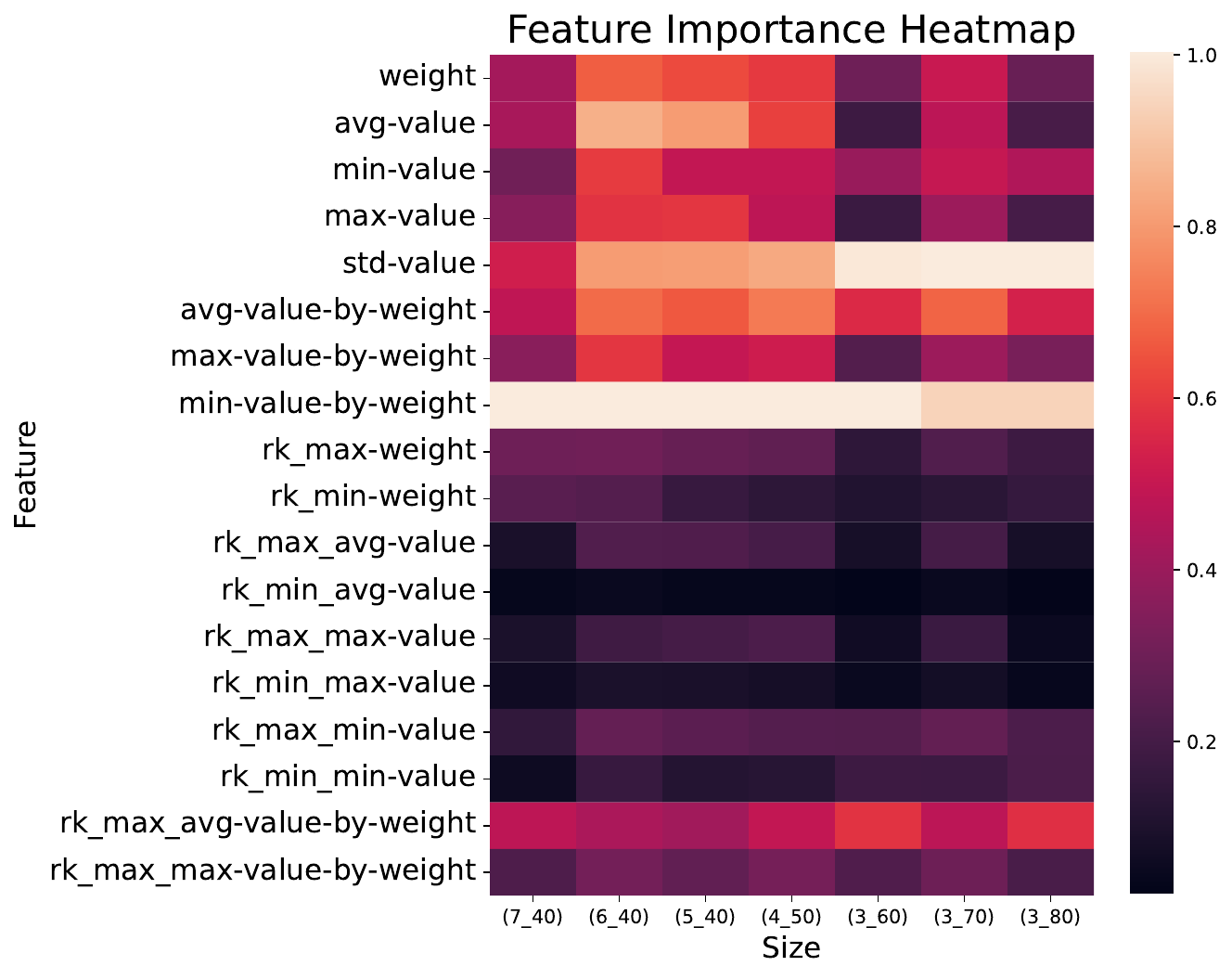}
    \caption{Heatmap of feature importance scores normalized across different sizes.     
    {The features are described in \Cref{tab:mkp_features}.
    The property-based features assume the same names as the properties detailed in \Cref{tab:prop_mkp}.
    The rank-based features of a variable have a prefix ``rk\_'' before the heuristic ordering; refer to \Cref{tab:ratio_of_sums} for more details.
    Finally, the ``Value SD'' feature from~\Cref{tab:mkp_features} is named std-value in this plot.}}
    \label{fig:feature_heatmap}
\end{figure}

\Cref{fig:feature_heatmap} is a heatmap of feature importance scores for different sizes. 
We can note that the min-value-by-weight feature is important across all sizes, especially for cases with more than 3 objectives. 
In fact, the choice of \vo{} heuristic \minValByWt{} was driven by feature importance scores and is a case in point of how learning-based methods can assist in designing good heuristics for optimization.
Furthermore, the real-valued features are more important than the categorical rank features for problems with more than 3 objectives.

For problems with 3 objectives, the std-value feature deviation is extremely crucial. 
Also, rank feature rk\_max\_avg-value-by-weight receives higher importance than some of the real-valued features.
It is also interesting to observe that certain rank-based features are consistently ignored across all sizes.

In a nutshell, the heatmap of feature importance scores helps in interpreting the features that govern the decisions made by \method{}, which also happens to be in alignment with a widely used heuristic to solve the single-objective knapsack problem.

\subsubsection{Q6. Can we train a single size-independent ML model?}
We answer in the affirmative: in fact, the methods ML+A and ML+AC in~\Cref{tab:knapsack_test_time} refer to two models trained on the union of all training datasets of different sizes. As described in the section ``Learning Models'', ML+A uses the same variable features as~\ml{}, whereas ML+AC adds the instance context features described in~\Cref{tab:mkp_features}. These two models perform remarkably well, sometimes even outperforming the size-specific~\ml{} models that are being tested in-distribution. 

The main takeaway here is that our ML model architecture, which is independent of the number of variables in an instance, enables the training of unified models that are size-independent. In turn, these models perform very well. Combined with the simple features and types of ML models we have used, this finding points to a great potential for the use of ML in multiobjective BDD problems. The appendix includes feature importance plots for both of these models. ML+AC seems to make great use of the context features in its predictions, as expected.



%% file: Sections/conclusion.tex
\section{Conclusion}
\label{sec:conclusion}

\method{} is the first machine learning framework for accelerating the BDD approach for multiobjective integer linear programming. We contribute a number of techniques and findings that may be of independent interest. For one, we have shown that variable ordering does impact the solution time of this approach. Our labeling method of using a black-box optimizer over a fixed-size parameter configuration space to discover high-quality variable orderings successfully bypasses the curse of dimensionality; the application of this approach to other similar labeling tasks may be possible, e.g., for backdoor set discovery in MIP~\cite{khalil2022finding}. An additional innovation is using size-independent ML models, i.e., models that do not depend on a fixed number of decision variables. This modeling choice enables the training of unified ML models, which our experiments reveal to perform very well.
Through a comprehensive case study of the knapsack problem, we show that~\method{} can produce variable orderings that significantly reduce \paretofront{} enumeration time.

There are several exciting directions for future work that can build on~\method{}:
\begin{enumerate}
    \item[--] The BDD approach to multiobjective integer programming has been applied to a few problems other than knapsack~\cite{bergman2021network}. Much of~\method{} can be directly extended to such other problems, assuming that their BDD construction is significantly influenced by the variable ordering. One barrier to such an extension is the availability of open-source BDD construction code. As the use of BDDs in optimization is a rather nascent area of research, it is not uncommon to consider a case study of a single combinatorial problem, as was done for example for Graph Coloring in~\cite{karahalios2022variable} and Maximum Independent Set in~\cite{bergman2012variable}.

    \item[--] Our method produces a~\textit{static} variable ordering upfront of BDD construction. While this was sufficient to improve on non-ML orderings for the knapsack problem, it may be interesting to consider~\textit{dynamic} variable orderings that observe the BDD construction process layer by layer and choose the next variable accordingly, as was done in~\cite{cappart2019improving}.

    \item[--] We have opted for rather interpretable ML model classes but the exploration of more sophisticated deep learning approaches may enable closing some of the remaining gap in training/validation loss, which may improve downstream solving performance.

    \item[--] Beyond the exact multiobjective setting, extending~\method{} to a heuristic that operates on a \textit{restricted} BDD may provide an approximate \paretofront{} much faster than full enumeration. We believe this to be an easy extension of our work. 
    
\end{enumerate}

%% file: Sections/appendix.tex
\subsection{Feature importance plots for size-independent ML models}
\begin{figure}[htbp!]
    \centering
    \includegraphics[width=\linewidth]{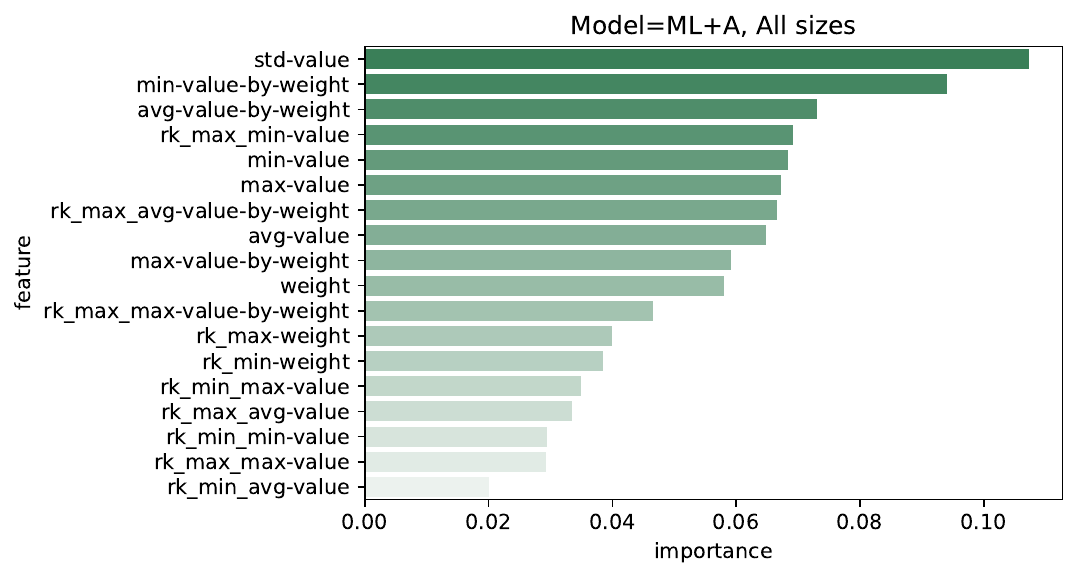}
    \caption{Feature importance plot for ML+A model. Refer to the caption of~\Cref{fig:feature_heatmap} for an explanation of the feature naming conventions.}
    \label{fig:feat_imp_all}
\end{figure}

\begin{figure}[htbp!]
    \centering
    \includegraphics[width=\linewidth]{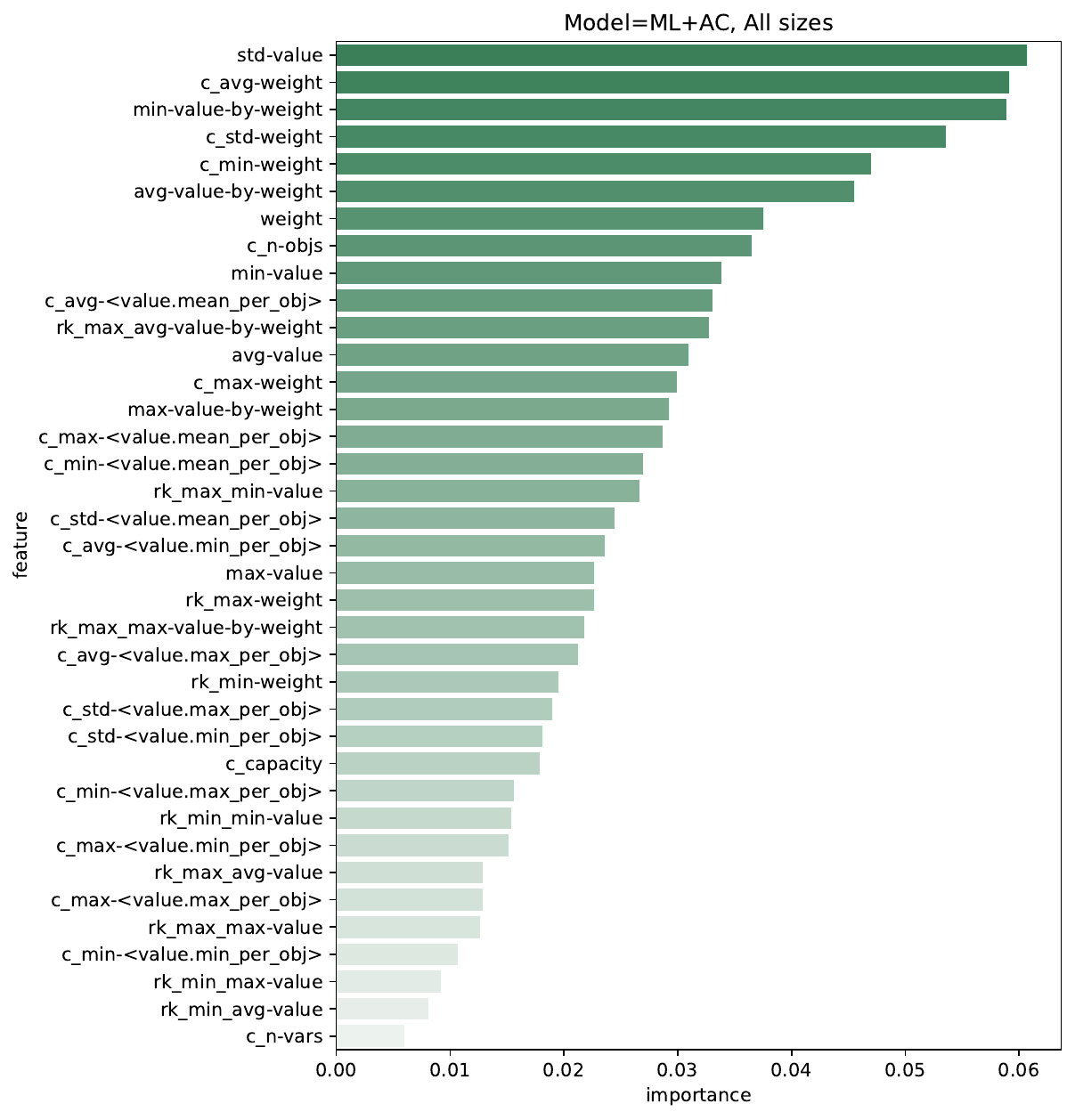}
    \caption{Feature importance plot for ML+AC model. Features that start with ``c\_'' are context features; see~\Cref{tab:mkp_features}.}
    \label{fig:feat_imp_all_context}
\end{figure}